\pdfoutput=1

\documentclass[11pt]{article}

\usepackage{naacl2021}

\usepackage{times}
\usepackage{latexsym}

\usepackage[T1]{fontenc}

\usepackage[utf8]{inputenc}

\usepackage{microtype}

\usepackage{todonotes}
\usepackage{multirow}
\usepackage{amsmath}
\usepackage{amsfonts}
\usepackage[capitalise]{cleveref}
\usepackage{csquotes}
\usepackage{colortbl}
\usepackage{comment}
\usepackage{arydshln}
\usepackage{caption}
\usepackage{subcaption}

\title{Attention vs non-attention for a Shapley-based explanation method}

\author{Tom Kersten \\
  University of Amsterdam \\
  \texttt{t.kersten@uva.nl} \\\And
  Hugh Mee Wong \\
  University of Amsterdam \\
  \texttt{h.m.wong@uva.nl} \\\AND
  Jaap Jumelet \\
  ILLC, University of Amsterdam \\
  \texttt{j.w.d.jumelet@uva.nl} \\\And
  Dieuwke Hupkes \\
  Facebook AI Research\\
  \texttt{dieuwkehupkes@fb.com} \\}

\begin{document}
\maketitle
\begin{abstract}
The field of explainable AI has recently seen an explosion in the number of explanation methods for highly non-linear deep neural networks. 
The extent to which such methods -- that are often proposed and tested in the domain of computer vision -- are appropriate to address the explainability challenges in NLP is yet relatively unexplored.
In this work, we consider \textit{Contextual Decomposition} (CD) -- a Shapley-based input feature attribution method that has been shown to work well for recurrent NLP models -- and we test the extent to which it is useful for models that contain attention operations. 
To this end, we extend CD to cover the operations necessary for attention-based models. 
We then compare how long distance subject-verb relationships are processed by models with and without attention, considering a number of different syntactic structures in two different languages: English and Dutch. 
Our experiments confirm that CD can successfully be applied for attention-based models as well, providing an alternative Shapley-based attribution method for modern neural networks.
In particular, using CD, we show that the English and Dutch models demonstrate similar processing behaviour, but that under the hood there are consistent differences between our attention and non-attention models.
\end{abstract}

\section{Introduction}
Machine learning models using deep neural architectures have seen tremendous performance improvements over the last few years. The advent of models such as LSTMs \cite{Hochreiter1997LongMemory} and, more recently, attention-based models such as Transformers \cite{Vaswani2017AttentionNeed} have allowed some language technologies to reach near human levels of performance. However, this performance has come at the cost of the interpretability of these models: high levels of non-linearity make it a near impossible task for a human to comprehend how these models operate.

Understanding how non-interpretable black box models make their predictions has become an active area of research in recent years \citep[i.a.]{hupkes2018visualisation, jumelet-hupkes-2018-language, samek2019explainable, ws-2019-2019-acl, tenney2019bert, ettinger2020bert}. 
One popular interpretability approach makes use of \textit{feature attribution methods}, that explain a model prediction in terms of the \textit{contributions} of the input features. 
For instance, a feature attribution method for a sentiment analysis task can tell the modeller how much each of the input words contributed to the decision of a particular sentence. 

Multiple methods of assigning contributions to the input feature approaches exist.
Some are based on local model approximations \cite{Ribeiro2016WhyClassifier}, others on gradient-based information \cite{DBLP:journals/corr/SimonyanVZ13, Sundararajan2017AxiomaticNetworks} and yet others consider perturbation-based methods \cite{Lundberg2017APredictions} that leverage concepts from game theory such as Shapley values \cite{shapley1953value}. 
Out of these approaches the Shapley-based attribution methods are computationally the most expensive, but they are better able at explaining more complex model dynamics involving feature interactions.
This makes these methods well-suited for explaining the behaviour of current NLP models on a more linguistic level.

In this work, we therefore focus our efforts on that last category of attribution methods, focusing in particular on a method known as Contextual Decomposition \cite[CD,][]{Murdoch2018BeyondLSTMs}, which provides a polynomial approach towards approximating Shapley values.
This method has been shown to work well on recurrent models without attention \citep{jumelet-etal-2019-analysing, saphra2020lstms}, but has not yet been used to provide insights into the linguistic capacities of attention-based models.
Here, to investigate the extent to which this method is also applicable for attention based models, we extend the method to include the operations required to deal with attention-based models and we compare two different recurrent models: a multi-layered LSTM model \citep[similar to][]{jumelet-etal-2019-analysing}, and a Single Headed Attention RNN \citep[SHA-RNN,][]{Merity2019SingleHead}.
We focus on the task of \emph{language modelling} and aim to discover simultaneously whether attribution methods like CD are applicable when attention is used, as well as how the attention mechanism influence the resulting feature attributions, focusing in particular on whether these attributions are in line with human intuitions.
Following, i.a.\ \citet{jumelet-etal-2019-analysing}, \citet{lakretz-etal-2019-emergence} and \citet{giulianelli-etal-2018-hood}, we focus on how the models process long-distance subject verb relationships across a number of different syntactic constructions.
To broaden our scope, we include two different languages: English and Dutch.

Through our experiments we find that, while both English and Dutch language models produce similar results, our attention and non-attention models behave differently. These differences manifest in incorrect attributions for the subjects in sentences with a plural subject-verb pair, where we find that a higher attribution is given to a plural subject when a singular verb is used compared to a singular subject.

Our main contributions to the field thus lie in two dimensions: on the one hand, we compare attention and non-attention models with regards to their explainability. On the other hand, we perform our analysis in two languages, namely Dutch and English, to see if patterns hold in different languages.

\section{Background}
In this section we first discuss the model architectures that we consider. Following this, we explain the attribution method that we use to explain the different models. 
Finally, we consider the task which we use to extract explanations. 

\subsection{Model architectures}
To examine the differences between attention and non-attention models, we look at one instance of each kind of model. 
For the attention model, we consider the Single Headed Attention RNN \cite[SHA-RNN,][]{Merity2019SingleHead}, and for our non-attention model a multi-layered LSTM \cite{Gulordava2018ColorlessHierarchically}. 
Since both models use an LSTM at their core, we hope to capture and isolate the influence of the attention mechanism on the behaviour of the model. 
Using a Transformer architecture instead would have made this comparison far more challenging, given that these kinds of models differ in multiple significant aspects from LSTMs with regards to their processing mechanism.
Below, we give a brief overview of the SHA-RNN architecture. 

\paragraph{SHA-RNN} The attention model we consider is the Single Headed Attention RNN, or SHA-RNN, proposed by \citet{Merity2019SingleHead}. 
The SHA-RNN was designed to be a reasonable alternative to the comparatively much larger Transformer models. \citeauthor{Merity2019SingleHead} argues that while larger models can bring better performance, this often comes at the cost of training and inference time. 
As such, the author proposed this smaller model, which achieves results comparable to earlier Transformer models, without hyperparameter tuning.

\begin{figure}
    \centering
    \includegraphics[width=\linewidth]{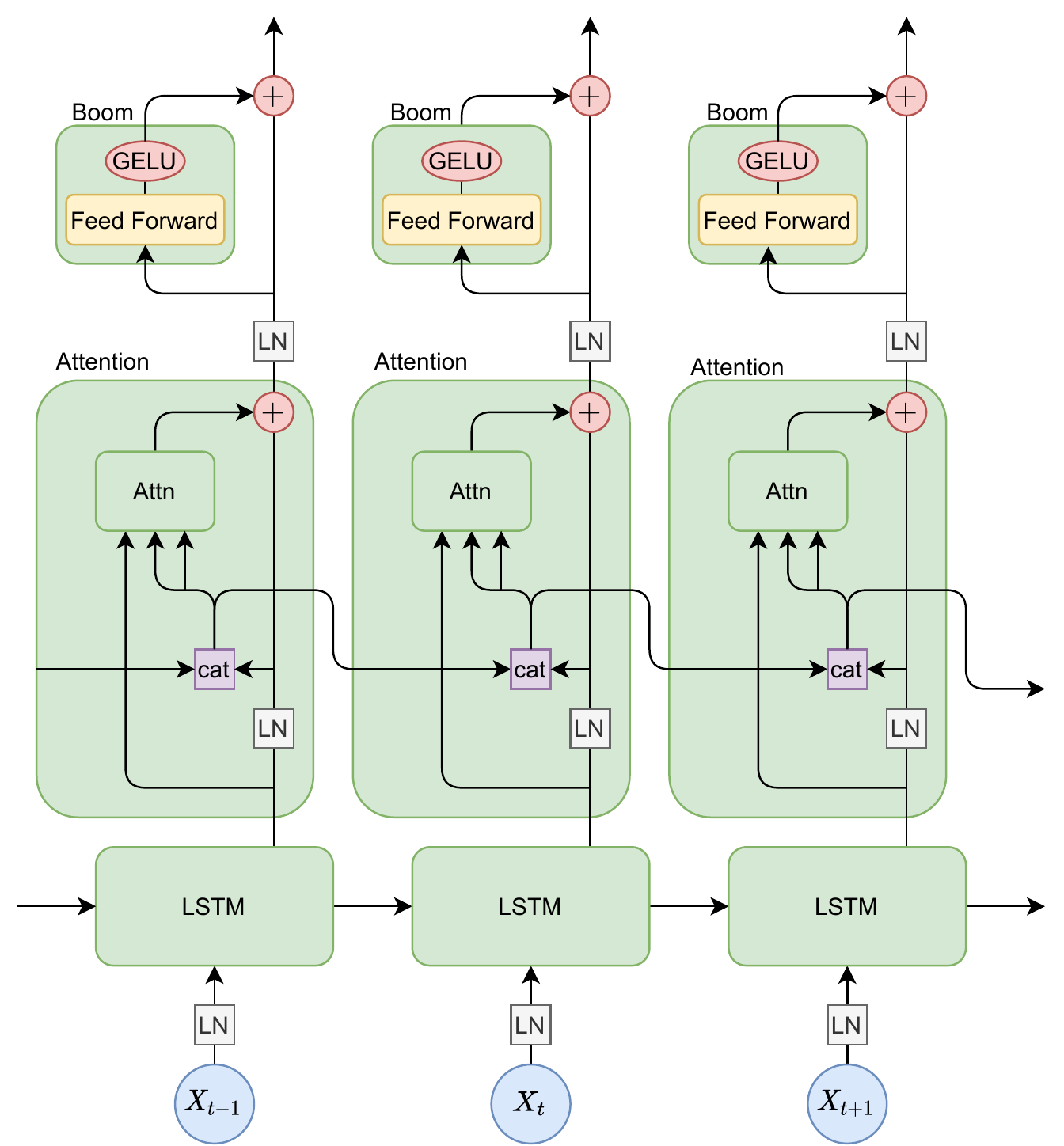}
    \caption{A schematic overview of a block in the SHA-RNN. A block in the SHA-RNN is composed of an LSTM, a single headed attention layer and a Boom feed-forward layer. Throughout the model, layer normalisation is used. Hidden states are passed between subsequent steps in the model. The memory state is concatenated with previous memory states, and passed on as well.}
    \label{fig:SHA-RNN}
\end{figure}

The SHA-RNN consists of a block structure with three modules: an LSTM, a pointer-based attention layer and a feed-forward Boom layer (we provide a graphical overview in Figure~\ref{fig:SHA-RNN}). These blocks can be stacked to create a similar setup to that of an encoder Transformer. Layer normalisation is applied at several points in the model.

The attention layer in the SHA-RNN uses only a single attention head, creating a similar mechanism to \citet{Grave2016ImprovingCache} and \citet{Merity2016PointerModels}. This is in contrast to most other Transformer (and thus attention) models, which utilise multiple attention heads. However, recent work, like \citet{Michel2019AreOne}, has shown that using only a single attention head may in some cases provide similar performance to a multi-headed approach, while significantly reducing the computational cost. Importantly, when using multiple blocks of the SHA-RNN, the attention layer is only applied in the second to last block.

The Boom layer represents the feed-forward layers commonly found in Transformer models \cite{Vaswani2017AttentionNeed}. In his work, \citeauthor{Merity2019SingleHead} uses a single feed-forward layer with a GELU activation \cite{Hendrycks2016GaussianGELUs}, followed by summation over the output to reduce the dimension of the resulting vector to that before applying the feed-forward layer.

\subsection{Contextual Decomposition}
The interpretability method that we use and extend in this paper is Contextual Decomposition  \cite[CD][]{Murdoch2018BeyondLSTMs}, a feature attribution method for explaining individual predictions made by an LSTM. 
CD decomposes the output into a sum of two contribution types $\beta+\gamma$: one part resulting from a specific ``relevant'' token or phrase ($\beta$), and one part resulting from all other input to the model ($\gamma$), which is said to be ``irrelevant''. 
The token or phrase of interest is provided as an additional parameter to the model.

CD performs a modified forward pass through the model for each individual token in the input sentence.
The $\beta+\gamma$ decomposition is achieved by splitting up the hidden and cell state of the LSTM into two parts as well:
\begin{align}
    h_t &= \beta_t + \gamma_t\\
    c_t &= \beta^c_t + \gamma^c_t
\end{align}

This decomposition is constructed such that $\beta$ corresponds to contributions made solely by elements in the relevant phrase, while $\gamma$ represents all other contributions.
Fundamental to CD is the role of interactions between $\beta$ and $\gamma$ terms that arrive from operations such as (point-wise) multiplications.
CD resolves this by ``factorizing'' the outcome of a non-linear activation function into a sum of components, based on an approximation of the Shapley value of the activation function \citep{shapley1953value}.

For example, the forget gate update of the cell state in an LSTM is defined as
\begin{equation}
    c_t' = c_{t-1} \odot \sigma(W_f x_t + V_f h_{t-1} + b_f)
\end{equation}
where $W_f \in \mathbb{R}^{d_x \times d_h}$, $V_f \in \mathbb{R}^{d_h \times d_h}$ and $b_f \in \mathbb{R}^{d_h}$. 
CD decomposes both $c_{t-1}$ and $h_{t-1}$ into a sum of $\beta$ and $\gamma$ terms:
\begin{align}
    c_t' &= (\beta^c_{t-1} + \gamma^c_{t-1}) \nonumber\\
    & \odot \sigma(W_f x_t + V_f (\beta_{t-1} + \gamma_{t-1}) + b_f)
\end{align}
The forget gate is then decomposed into a sum of four components ($x, \beta, \gamma~\&~b_f)$, based on their Shapley values, which leads to a cross product between the terms in the decomposed cell state, and the decomposed forget gate.
The $\beta+\gamma$ decomposition of the new cell state $c_t$ is formed by determining which specific interactions between $\beta$ and $\gamma$ components should be assigned to the new $\beta^c_t$ and $\gamma^c_t$ terms.

In this work, we consider the generalisation of the CD method proposed by \citet{jumelet-etal-2019-analysing}, namely Generalized Contextual Decomposition (GCD). They alter the way that $\beta$ and $\gamma$ interactions are divided over these terms. As such, this method provides a more complete picture of the interactions within the model. 
For a more detailed explanation of the procedure we refer to the original papers.

\subsection{Number Agreement Task}
To test our models, we consider the Number Agreement (NA) task, a linguistic task that has stood central in various works in the interpretability literature \citep{lakretz-etal-2019-emergence, Linzen2016AssessingDependencies, Gulordava2018ColorlessHierarchically, wolf2019, goldberga}. 
In this task, a model is evaluated by how well it is able to track the subject-verb relations over long distances, as assessed by the percentage of cases in which the model is able to match the form of the verb to the number of the subject. 
The challenge in the NA task lies in the presence of one or more attractor nouns between the subject and the verb that competes with the subject. 
For instance in the sentence "The boys at the car greet", "car" forms the attractor noun, and is a different number than the boys, thereby possibly confusing the model to predict a singular verb, "greets". 
 
Several earlier studies preceded us in considering number agreement as a means to investigate language models. \citeauthor{Linzen2016AssessingDependencies} laid the groundwork for this task, using it to assess the ability of LSTMs to learn syntax-sensitive dependencies. In their work, they only considered the English language. \citet{Gulordava2018ColorlessHierarchically} extended the task to the Italian, Hebrew and Russian languages. Moreover, they provided a more in-depth study of the Italian model, comparing it to human subjects. \citet{lakretz-etal-2019-emergence} provided a detailed look at the underlying mechanisms of LSTMs by which they are able to model grammatical structure. To this end, they performed an ablation study and discovered which units were mainly responsible for this mechanism. Finally, further research into the Italian version of the NA task in \citet{DBLP:journals/corr/abs-2006-11098} investigated how emergent mechanisms in  language models relate to linguistic processing in humans.

Number agreement has also been explored before in the context of attribution methods.
Due to the clear dependency between a subject and a verb, it is a useful task to evaluate whether a model based its prediction of the verb on the number information of the subject.
\citet{poerner2018evaluating} provide a large suite of evaluation tasks for attribution methods including number agreement, and show that attribution methods can sometimes yield unexpected contribution patterns.
\citet{jumelet-etal-2019-analysing} employ Contextual Decomposition to investigate the behaviour of an LSTM LM on a number agreement task, and demonstrate that their model employs a \textit{default reasoning} heuristic when resolving the task, with a strong bias for singular verbs.
\citet{DBLP:journals/corr/abs-2005-00062} investigates an attribution method on a range of number agreement constructions containing relative clauses, showing that LMs possess a robust notion of number information.

\section{Method}
In this section, we first look at extending Contextual Decomposition for the SHA-RNN. Following this, we outline the models which we will use for our experiments. Finally, we explain how we extended the Number Agreement task and how we applied Contextual Decomposition to the NA task, forming the Subject Attribution task.

\subsection{Contextual Decomposition for the SHA-RNN}
The original Contextual Decomposition paper \cite{Murdoch2018BeyondLSTMs} only defines the decomposition for an LSTM model. 
The SHA-RNN also contains several operations that have not previously been covered by these two papers. 
As such, we have defined the decompositions for the following two operations: Layer Normalization \cite{Ba2016LayerNormalization} and the Softmax operation in the Single Headed Attention layer \cite{Merity2019SingleHead}. 
Based on these new decompositions, we leverage the implementation of Contextual Decomposition in the \mbox{\texttt{diagNNose}} library of \citet{Jumelet2020DiagNNose} to also cover our SHA-RNN.

\paragraph{Layer Normalization}
Layer Normalization estimates the normalization statistics over the summed inputs to the neurons in a hidden layer. A definition of the Layer Normalization operation can be found in \cref{eq:ln}.  
\begin{equation}
    \begin{split}
    \mu &= \frac{1}{n} \sum_{i=1}^n a_i,\\
    \sigma &= \sqrt{\frac{1}{n} \sum_{i=1}^n (a_i - \mu)^2},\\
    \text{LN}(a) &= \alpha \frac{a - \mu}{\sigma} + \delta,
    \end{split}
    \label{eq:ln}
\end{equation}
where $a$ represents the inputs to the hidden layer, $n$ the number of hidden units and $\alpha$ and $\delta$ are learnable parameters.

Because it looks at all inputs in a layer, both $\beta$ and $\gamma$ might interact within this layer. As such, we must define how we handle the decomposition of this operation, which we show in \cref{eq:ln-decomp}.
\begin{equation}
    \begin{split}
        \beta^{l+1} &= \text{LN}(\beta^l) - \delta,\\
        \gamma^{l+1} &= \text{LN}(\beta^l + \gamma^l) - \text{LN}(\beta^l) + \delta\\
        \text{LN}(a) &= \text{LN}(\beta^l + \gamma^l) = \beta^{l+1} + \gamma^{l+1}
    \end{split}
    \label{eq:ln-decomp}
\end{equation}
Our decomposition strictly separates the $\gamma$ contributions from the $\beta$ contributions, which means that no information from $\gamma$ may be captured in $\beta$.

\paragraph{Softmax}
Similar to our treatment of the Layer Normalization operation, we strictly separate $\gamma$ from the $\beta$ components, as can be observed in \cref{eq:sm-decomp}.
\begin{equation}
    \begin{split}
        \beta^{l+1} &= \text{Softmax}(\beta^l),\\
        \gamma^{l+1} &= \text{Softmax}(\beta^l + \gamma^l) - \beta^{l+1}
    \end{split}
    \label{eq:sm-decomp}
\end{equation}

\subsection{Models}
For our experiments we consider two types of models: the attention SHA-RNN model and the non-attention LSTM model. Below, we will outline the specific architectures used and training hyperparameters chosen to build and train these models.

\subsubsection{Architectures}

\paragraph{LSTM model}
The LSTM model we use is similar to the one used by \citet{Gulordava2018ColorlessHierarchically}. The model is a stacked two layer LSTM, each with 650 hidden units. Word embeddings are trained alongside the model and the weights of the embedding layer are tied to the decoder layer \cite{Inan2016TyingModeling}.

\paragraph{SHA-RNN model}
For our SHA-RNN we use two blocks (see \cref{fig:SHA-RNN}), each with an LSTM with 650 hidden units. Furthermore, our model also utilises a trained word embedding layer with tied weights, similar to our non-attention model. Finally, our Boom layer does not increase our dimension size, but keeps it at 650. This means our Boom layer reduces to a feed-forward layer with GELU activations.

\subsubsection{Training}
We trained four models to conduct our experiments on. For both the attention (SHA-RNN) and non-attention (LSTM) model architectures, a model was trained on a Dutch and English corpus. Both corpora are based on wikipedia text. Following \citet{Gulordava2018ColorlessHierarchically}, only the 50.000 most common words were retained in the vocabulary for both corpora, replacing all other words with \texttt{<unk>} tokens. The corpora were split into a training, validation and test set.

The training of the models is split up in two phases: first, the model is trained for thirty epochs with a learning rate of $0.02$ and a batch size of 64. Then, we fine-tune the model for an additional five epochs with the learning rate halved to $0.01$ and a batch size of 16. During training, we set dropout to $0.1$. We use the LAMB optimizer \cite{DBLP:journals/corr/abs-1904-00962} following \citet{Merity2019SingleHead}.

\begin{table*}[t]
    \centering
    \resizebox{\linewidth}{!}{
\begin{tabular}{c|l|l}
    \textbf{NA-task} & \textbf{Template} & \textbf{Example} \\
    \hline
    Simple & \textsc{det \underline{n} v} & De \underline{jongen} groet\\
    & & The \underline{boy} greets \\
    \hdashline
    NounPP & \textsc{det \underline{n} prep det n v} & De \underline{jongens} bij de auto groeten\\
    & & The \underline{boys} at the car greet\\
    \hdashline
    NamePP & \textsc{det \underline{n} prep name v} & De \underline{jongens} bij Pat groeten\\
    & & The \underline{boys} at Pat greet\\
    \hdashline
    SConj & \textsc{det n v} en/and \textsc{det \underline{n} v} & De jongen groet en de \underline{moeders} missen\\
    & & The boy greets and the \underline{mothers} miss\\
    \hdashline
    ThatNounPP & \textsc{det n v} dat/that \textsc{det \underline{n} prep det n v} & De jongen denkt dat de \underline{moeders} bij de auto missen\\
    & & The boy thinks that the \underline{mothers} at the car miss\\
\end{tabular}}

    \caption{Overview of the templates for the NA-tasks. \textsc{det} is a determiner, \textsc{n} a noun, \textsc{name} a name of a person, \textsc{v} a verb and \textsc{prep} a preposition. The underlined noun in the template signifies the subject belonging to the relevant verb.}
    \label{tab:NA-tasks}
\end{table*}

\subsection{Extending Number Agreement}
In this work, we extend the Number Agreement (NA) task to the Dutch language. We do so by applying the same procedure that was used in \citet{lakretz-etal-2019-emergence}, namely by creating a synthetic dataset. This is different from the works of \citet{Linzen2016AssessingDependencies} and \citet{Gulordava2018ColorlessHierarchically}, which derived their sentences directly from corpora.

Our version of the NA task contains a total of five different templates. First of all, we use a simple template called Simple in which the verb immediately follows the subject. We then extend this by adding a prepositional phrase which modifies the subject between the subject and the verb, either by having a prepositional phrase containing a noun (NounPP) or containing a proper noun (NamePP). We then have the sentence conjunction (SConj) task, which consists of two Simple templates separated by a conjunction. The challenge of the SConj task is correctly predicting the number of the verb in the second sentence. Finally, we have the ThatNounPP template, which contains a declarative content clause which incorporates a second subject-verb dependency with a noun modifying prepositional phrase in its that-clause. An overview of the templates including example sentences can be found in \cref{tab:NA-tasks}.

We create our final NA-task by obtaining frequent words from our corpus to populate these sentence templates. This process is done for both the Dutch and the English corpora, such that we can more easily compare the results. 

\subsection{Subject Attribution Task}
We propose a new task for input feature attribution methods based on the Number Agreement task: Subject Attribution. The goal of the task to produce explanations in such a way that congruent subject-verb relations gain higher attributions than non-congruent ones. 

\begin{figure*}
    \centering
    \includegraphics[width=0.95\textwidth]{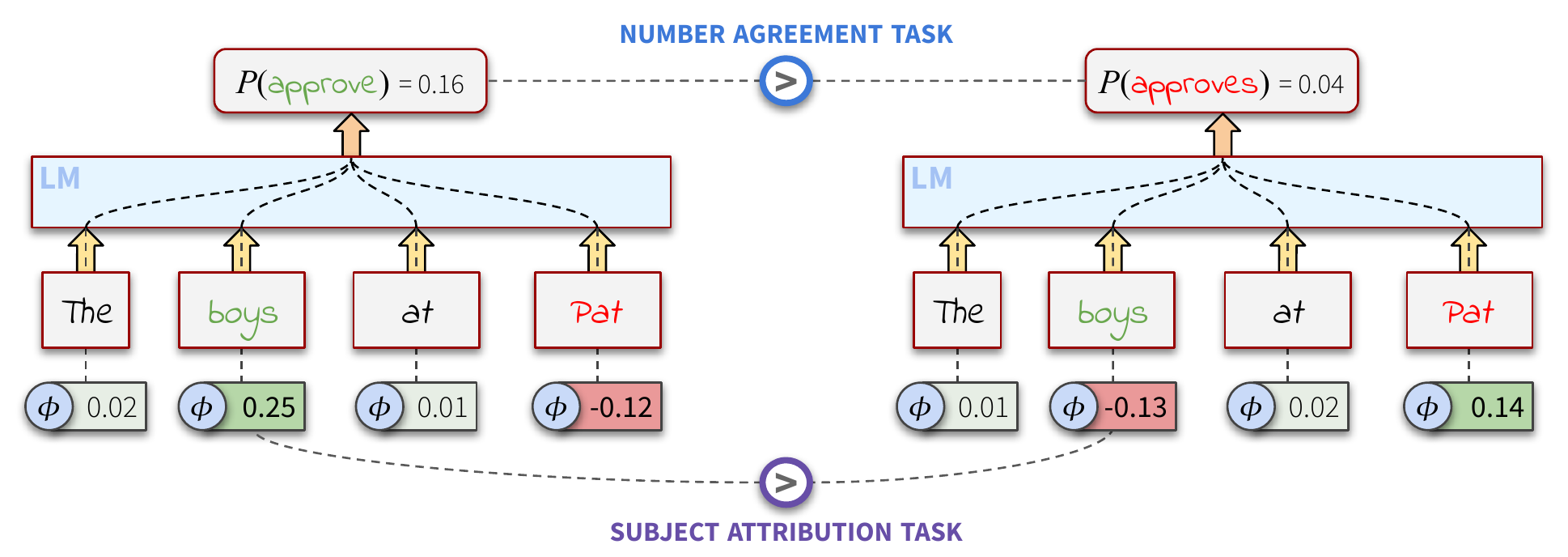}
    \caption{Schematic overview of the default \textbf{number agreement task} that compares the output probabilities of the LM, and the \textbf{subject attribution task} that compares the attribution scores of the subject to the correct and incorrect form of the verb.
    We hypothesise that for a model with a sophisticated understanding of number agreement, the subject's contribution to the correct verb form is greater than to the incorrect form. 
    }
    \label{fig:attribution_viz}
\end{figure*}

In context of the NA task this means that we compare the attribution scores of the subject of the sentence in the case where it is and is not congruent with the number of the verb. In our evaluation we consider a higher attribution for the congruent noun compared to the non-congruent noun to be correct, as this would be in line with human intuition. A schematic overview of this task can be found in \cref{fig:attribution_viz}.

In this work, we use the task in the following way: we apply our attribution method on each sentence within our dataset, generating input feature attributions. We then compare the subject attributions of these sentences to find in which percentage of the sentences the attributions for the subject were higher for the congruent verb than the non congruent one.

\section{Results and analysis}
In our work, we have considered several experiments. Firstly, we evaluate the ability of our models to handle the data itself by comparing the model perplexities. Following this, we look at the Number Agreement and Subject Attribution tasks to evaluate the differences between our models.

\subsection{Model Perplexities}
To establish the adequacy of our models on the data, we calculate the perplexity for each model over the held-out test set (\cref{tab:perplexity}). Due to the different data sets used for the two languages, direct comparisons between the perplexity scores for the English and Dutch models are not feasible. We do observe that for both languages, the SHA-RNN yields a perplexity score that is 5\% lower than the score of the LSTM counterpart.

\begin{table}
    \centering
    \begin{tabular}{c|c}
    Model & Perplexity \\
    \hline
    LSTM (English) & 56.24\\
    LSTM (Dutch) & 34.24\\
    SHA-RNN (English) & 53.25\\
    SHA-RNN (Dutch) & 32.54\\
\end{tabular}

    \caption{Model perplexities}
    \label{tab:perplexity}
\end{table}

\subsection{Number Agreement}
\label{sec:results_explanations}
To assess the performance of the different language models, we consider the different sentence structures presented in \cref{tab:NA-tasks}. For each sentence structure, we evaluate the predictive performance of the model on matching the form of the verb to the number of the relevant subject. For example, given a singular subject, we evaluate $p(\textsc{verb}_\textsc{S} | \textsc{subj}_\textsc{S}) > p(\textsc{verb}_\textsc{P} | \textsc{subj}_\textsc{S})$. The same sentence templates have been used for the Subject Attribution task. We apply Contextual Decomposition to the sentences to investigate the behavioural differences between the models. 

We examine the results of our experiments along two axes: language and attention. First, we compare the Dutch and English language models. Following this, we analyse the differences between the attention and non-attention models.

\begin{table*}[ht]
    \centering
    \begin{subfigure}{\linewidth}
        \centering
\scalebox{0.9}{
\begin{tabular}{lcccccc}
    \hline
    \multicolumn{1}{l}{\multirow{2}{*}{NA-task}} & \multicolumn{3}{c}{Singular Subject} & \multicolumn{3}{c}{Plural Subject} \\
                                                 & Condition    & SHA-RNN    & LSTM      & Condition    & SHA-RNN   & LSTM     \\
    \hline
     Simple      & \underline{S}             & \cellcolor[HTML]{8ac34f} {92.1} (77.8)        & \cellcolor[HTML]{cdeaa7} {99.2} (65.4)   & \underline{P}             &        \cellcolor[HTML]{e99cc8} {94.0} (25.9) &    \cellcolor[HTML]{e8f5d5} {94.4} (58.6)\\
     \hline
     NounPP      & \underline{S}S            & \cellcolor[HTML]{6dad36} {99.0} (83.3)       & \cellcolor[HTML]{edf6df} {94.7} (56.1)   & \underline{P}S            &        \cellcolor[HTML]{df79b0} {91.5} (20.7) &    \cellcolor[HTML]{b7e085} {98.5} (70.0)\\
     NounPP      & \underline{S}P            & \cellcolor[HTML]{75b43b} {95.2} (82.0)       & \cellcolor[HTML]{f8f3f6} {94.7} (48.3)   & \underline{P}P            &        \cellcolor[HTML]{e07eb3} {96.8} (21.3) &    \cellcolor[HTML]{b0dc7d} {98.7} (71.2) \\
     \hline
     NamePP      & \underline{S}             & \cellcolor[HTML]{e9f5d6} {59.3} (58.3)      & \cellcolor[HTML]{ebf6db} {81.8} (57.2)   & \underline{P}             &        \cellcolor[HTML]{fbe7f2} {83.8} (43.3)&    \cellcolor[HTML]{f8f4f6} {75.3} (48.8) \\
     \hline
     SConj      & S\underline{S}            & \cellcolor[HTML]{8fc654} {95.8} (77.0)       & \cellcolor[HTML]{4b8f21} {96.0} (90.3)   & S\underline{P}            &        \cellcolor[HTML]{fbe9f2} {88.7} (43.8) &    \cellcolor[HTML]{d8efb9} {89.3} (63.0) \\
     SConj      & P\underline{S}            & \cellcolor{green!25} {42.8}  (67.0)      & \cellcolor[HTML]{509423} {89.5} (89.3)   & P\underline{P}            &        \cellcolor[HTML]{f7f7f6} {94.0} (50.2) &    \cellcolor[HTML]{fbe6f1} {95.5} (42.8) \\
     \hline
     ThatNounPP & S\underline{S}S           & \cellcolor[HTML]{acd977} {98.3} (72.2)      & \cellcolor[HTML]{7bb93e} {96.7} (80.7)    & S\underline{P}S           &        \cellcolor[HTML]{ddf1c1} {99.3} (61.8) &   \cellcolor[HTML]{509423} {100.0} (89.3) \\
     ThatNounPP & S\underline{S}P           & \cellcolor[HTML]{cdeaa7} {99.0} (65.5)      & \cellcolor[HTML]{9acd61} {94.7} (75.2)   & S\underline{P}P           &        \cellcolor[HTML]{c9e8a2} {99.2} (66.2) &   \cellcolor[HTML]{45881f} {100.0} (91.8) \\
     ThatNounPP & P\underline{S}S           & \cellcolor[HTML]{b5df82} {97.8} (70.7)       & \cellcolor[HTML]{6dad36} {96.8} (83.5)   & P\underline{P}S           &        \cellcolor[HTML]{dbf0bf} {99.7} (62.2) &   \cellcolor[HTML]{509423} {100.0} (89.3) \\
     ThatNounPP & P\underline{S}P           & \cellcolor[HTML]{ddf1c1} {98.2} (62.0)       & \cellcolor[HTML]{8ac34f} {91.3} (78.0)   & P\underline{P}P           &        \cellcolor[HTML]{cbe9a4} {99.5} (65.8) &   \cellcolor[HTML]{468a20} {100.0} (91.7)\\
     \hline
\end{tabular}}

        \caption{Results for the \textbf{Dutch} language models.}
        \label{tab:pred_exp_dutch}
    \end{subfigure}\\[10pt]
    \begin{subfigure}{\linewidth}
        \centering
\scalebox{0.9}{
\begin{tabular}{lcccccc}
    \hline
    \multicolumn{1}{l}{\multirow{2}{*}{NA-task}} & \multicolumn{3}{c}{Singular Subject} & \multicolumn{3}{c}{Plural Subject} \\
                                                 & Condition    & SHA-RNN    & LSTM      & Condition    & SHA-RNN   & LSTM     \\
    \hline
    Simple      & \underline{S}             & \cellcolor[HTML]{54aa27} 94.0 (93.3)       & \cellcolor[HTML]{54aa27} 92.7 (93.3)   & \underline{P}             &        \cellcolor[HTML]{ff34ab} 99.3 (11.7) &   \cellcolor[HTML]{f8cee6} 96.3 (35.7)\\
    \hline
    NounPP      & \underline{S}S            & \cellcolor[HTML]{58af28} 86.0 (92.3)       & \cellcolor[HTML]{489c25} 78.3 (95.5)   & \underline{P}S            &        \cellcolor[HTML]{f61e9b} 82.5 (8.8) &    \cellcolor[HTML]{eff6e5} 93.3 (54.6) \\
    NounPP      & \underline{S}P            & \cellcolor[HTML]{52a827} 83.8 (93.5)      & \cellcolor[HTML]{50a527} 54.8 (94.0)  & \underline{P}P            &        \cellcolor[HTML]{f9209d} 97.0 (9.0) &    \cellcolor[HTML]{e7f5d2} 96.8 (59.5)\\
    \hline
    NamePP      & \underline{S}             & \cellcolor[HTML]{68c12d} 68.0 (89.3)       & \cellcolor[HTML]{429524} 86.7 (96.5)   & \underline{P}             &        \cellcolor[HTML]{ff59bf} 66.2 (14.5) &    \cellcolor[HTML]{ff42b3} 52.3 (12.5) \\
    \hline
    SConj      & S\underline{S}            & \cellcolor[HTML]{54aa27} 93.8 (93.0)       & \cellcolor[HTML]{61ba2a} 94.3 (90.5)   & S\underline{P}            &        \cellcolor[HTML]{ff67c7} 99.3 (15.7) &    \cellcolor[HTML]{75cc36} 96.3 (87.2) \\
    SConj      & P\underline{S}            & \cellcolor[HTML]{52a827} 82.3 (93.5)      & \cellcolor[HTML]{4ea326} 94.3 (94.3)   & P\underline{P}            &        \cellcolor[HTML]{ff2aa6} 99.3 (10.7) &    \cellcolor[HTML]{61ba2a} 98.8 (90.5)\\
    \hline
    ThatNounPP & S\underline{S}S           & \cellcolor[HTML]{338220} 91.8 (100.0)      & \cellcolor[HTML]{59b129} 70.7 (92.0)   & S\underline{P}S           &        \cellcolor[HTML]{da1185} 92.3 (5.0) &     \cellcolor[HTML]{f5f7f2} 95.8 (51.3)\\
    ThatNounPP & S\underline{S}P           & \cellcolor[HTML]{338220} 85.2 (100.0)     & \cellcolor[HTML]{50a527} 43.7 (94.0)   & S\underline{P}P           &        \cellcolor[HTML]{d50f81} 98.7 (4.2) &   \cellcolor[HTML]{cbe9a4} 100.0 (65.7) \\
    ThatNounPP & P\underline{S}S           & \cellcolor[HTML]{338220} 86.2 (99.8)       & \cellcolor[HTML]{58af28} 69.7 (92.3)   & P\underline{P}S           &        \cellcolor[HTML]{d71083} 92.0 (4.3) &    \cellcolor[HTML]{edf6e1} 97.0 (55.7)\\
    ThatNounPP & P\underline{S}P           & \cellcolor[HTML]{338220} 81.2 (100.0)       & \cellcolor[HTML]{58af28} 46.3 (92.3)   & P\underline{P}P           &        \cellcolor[HTML]{c70876} 98.2 (2.3) &    \cellcolor[HTML]{c0e593} 99.5 (68.0)\\
    \hline
\end{tabular}}

        \caption{Results for the \textbf{English} language models.}
        \label{tab:pred_exp_eng}
    \end{subfigure}
    \caption{Overview of prediction accuracy scores (the numbers outside the brackets) and subject attribution behaviour (in brackets) on the Number Agreement tasks for the Dutch and English language models. For each task, the noun inflections are given in the condition column, with S indicating singular and P indicating plural. The underlined letter in the condition indicates the noun belonging to the verb that is predicted. The numbers in brackets denote the performance on the subject attribution task: the percentage of cases in which the attributions of the subjects were higher to the congruent verb than to the non-congruent ones.
    The colour coding of the table cells follows the performance on this subject attribution task along a colour gradient from green (high performance) to red (low performance).
    }
    \label{tab:pred_exp}
\end{table*}

\begin{figure*}
    \centering
    \begin{subfigure}{0.33\linewidth}
        \centering
        \begin{subfigure}{0.45\linewidth}
            \centering
            \includegraphics[width=\linewidth]{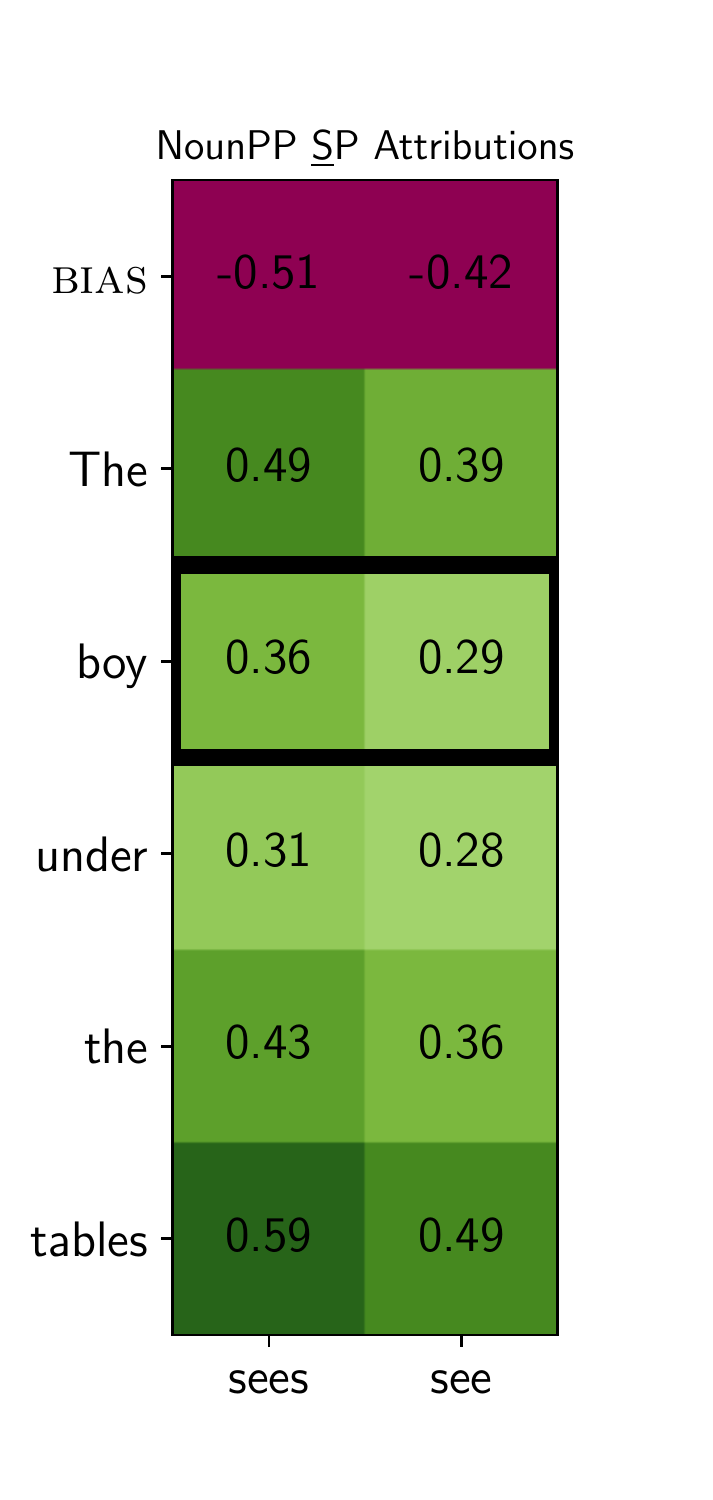}
            \label{fig:nounpp_sp_sharnn_single}
        \end{subfigure}%
        \begin{subfigure}{0.45\linewidth}
            \centering
            \includegraphics[width=\linewidth]{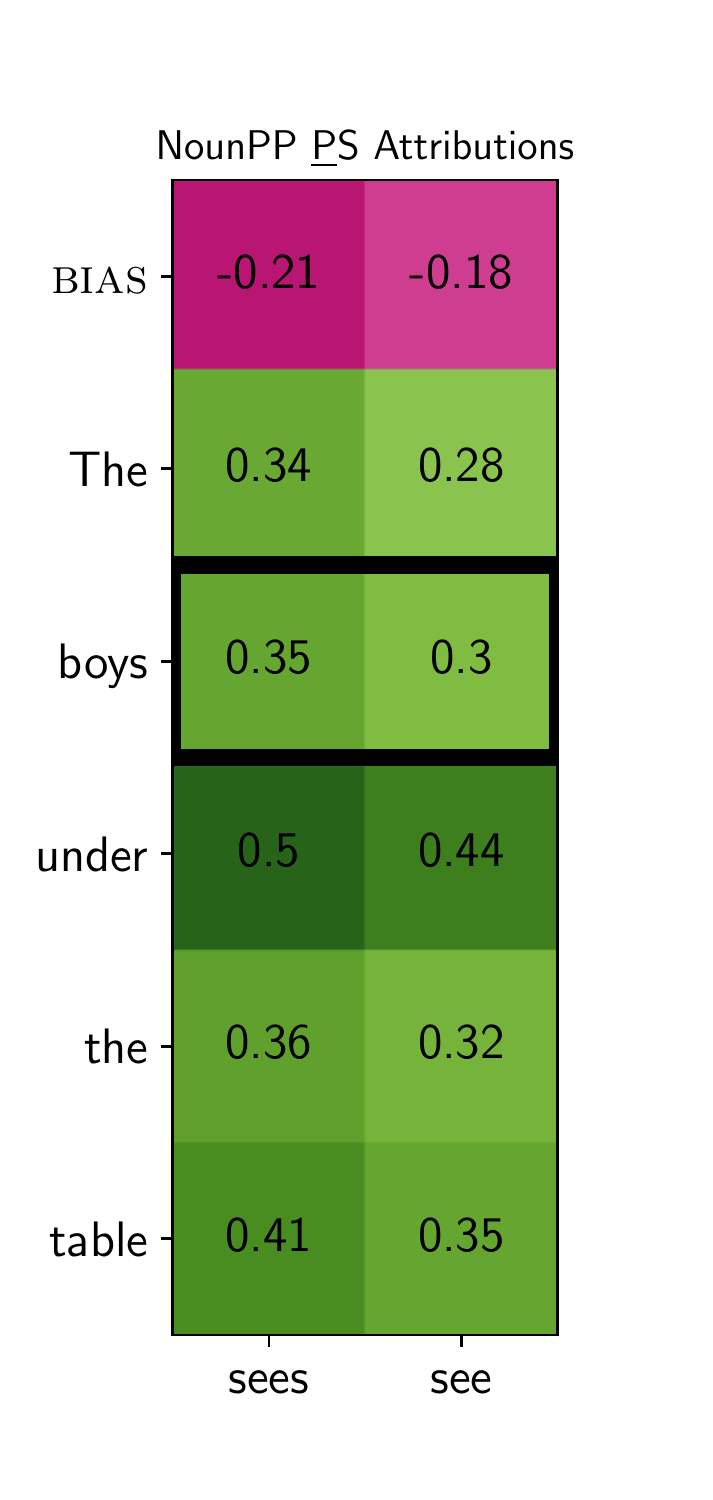}
            \label{fig:nounpp_ps_sharnn_single}
        \end{subfigure}%
        \caption{Example SHA-RNN attributions}
        \label{fig:nounpp_sharnn_single}
    \end{subfigure}%
    \begin{subfigure}{0.33\linewidth}
        \centering
        \begin{subfigure}{0.45\linewidth}
            \centering
            \includegraphics[width=\linewidth]{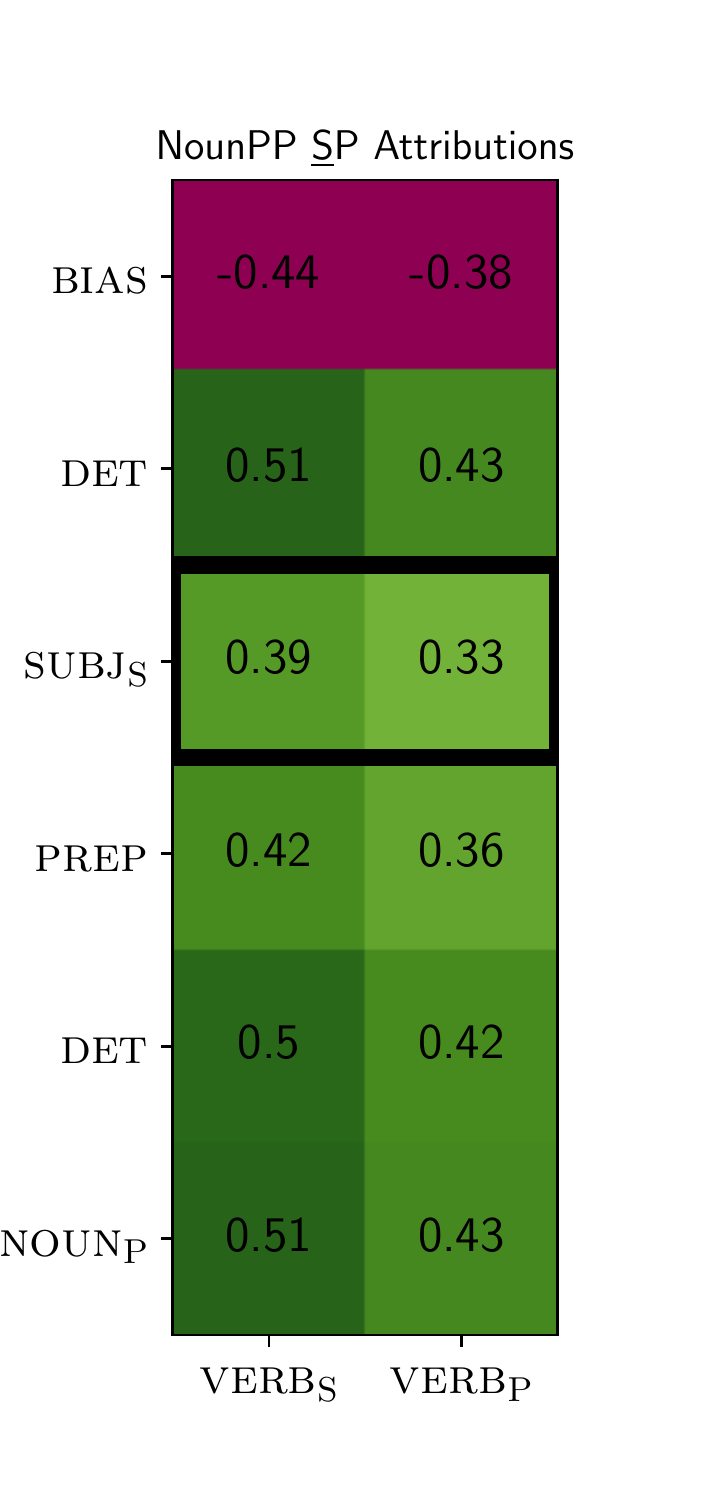}
            \label{fig:nounpp_sp_sharnn}
        \end{subfigure}%
        \begin{subfigure}{0.45\linewidth}
            \centering
            \includegraphics[width=\linewidth]{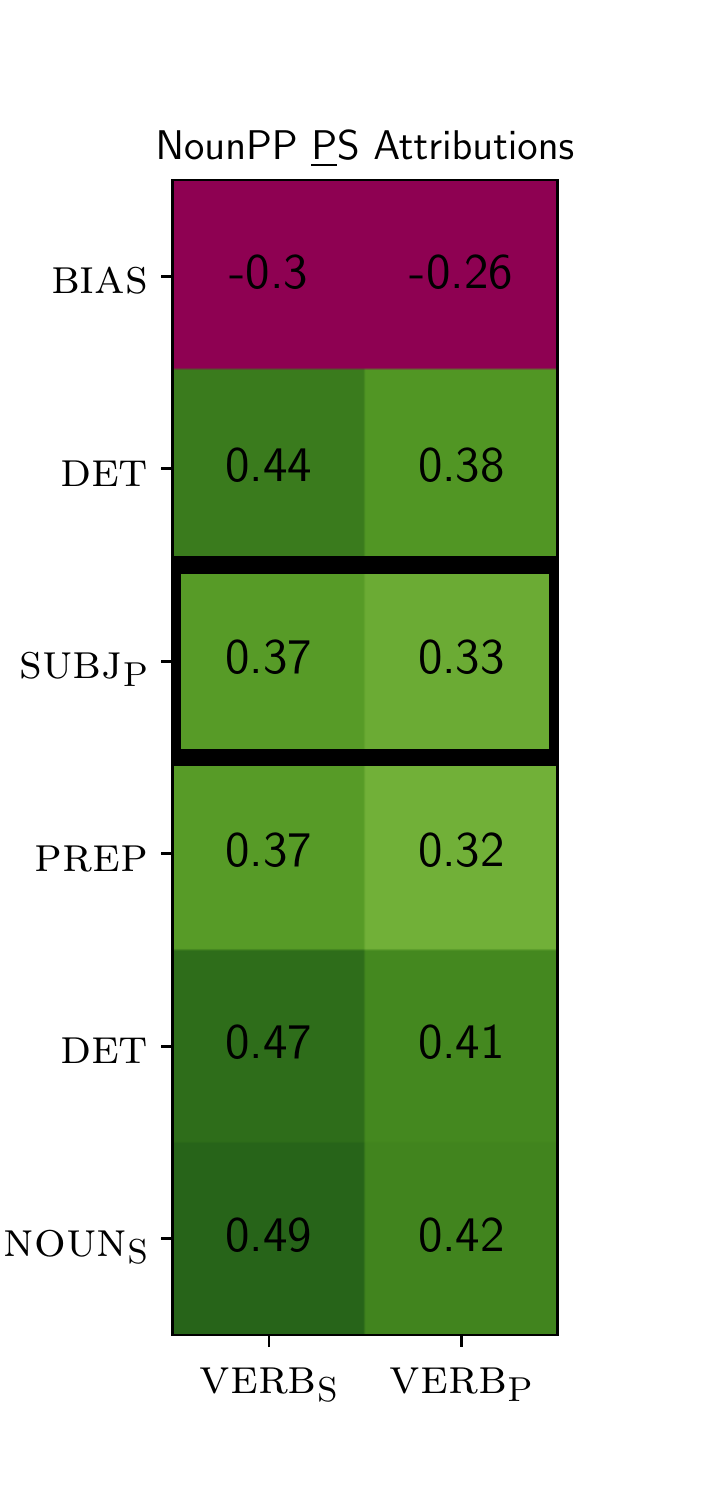}
            \label{fig:nounpp_ps_sharnn}
        \end{subfigure}
        \caption{Aggregated SHA-RNN attributions}
        \label{fig:nounpp_sharnn}
    \end{subfigure}%
    \begin{subfigure}{0.33\linewidth}
        \centering
        \begin{subfigure}{0.45\linewidth}
            \centering
            \includegraphics[width=\linewidth]{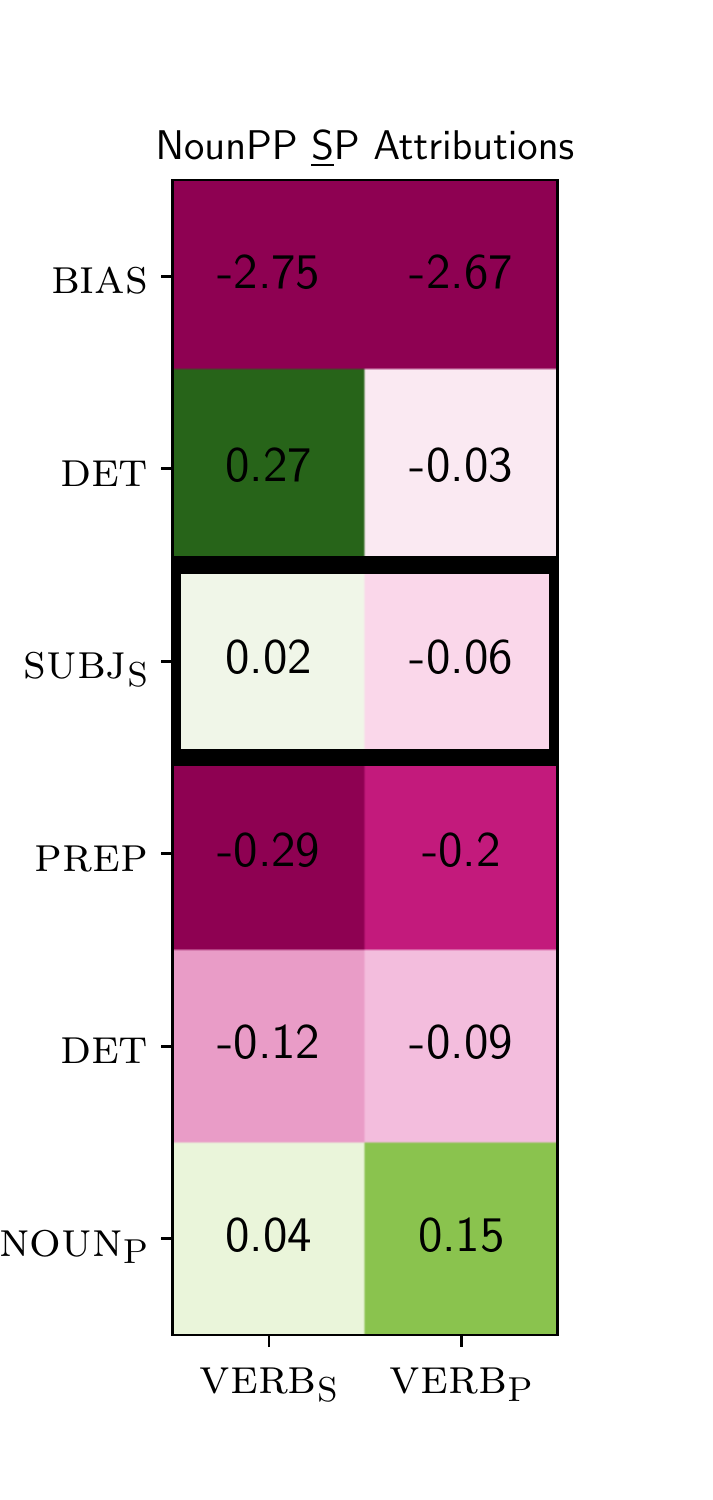}
            \label{fig:nounpp_sp_rnn}
        \end{subfigure}%
        \begin{subfigure}{0.45\linewidth}
            \centering
            \includegraphics[width=\linewidth]{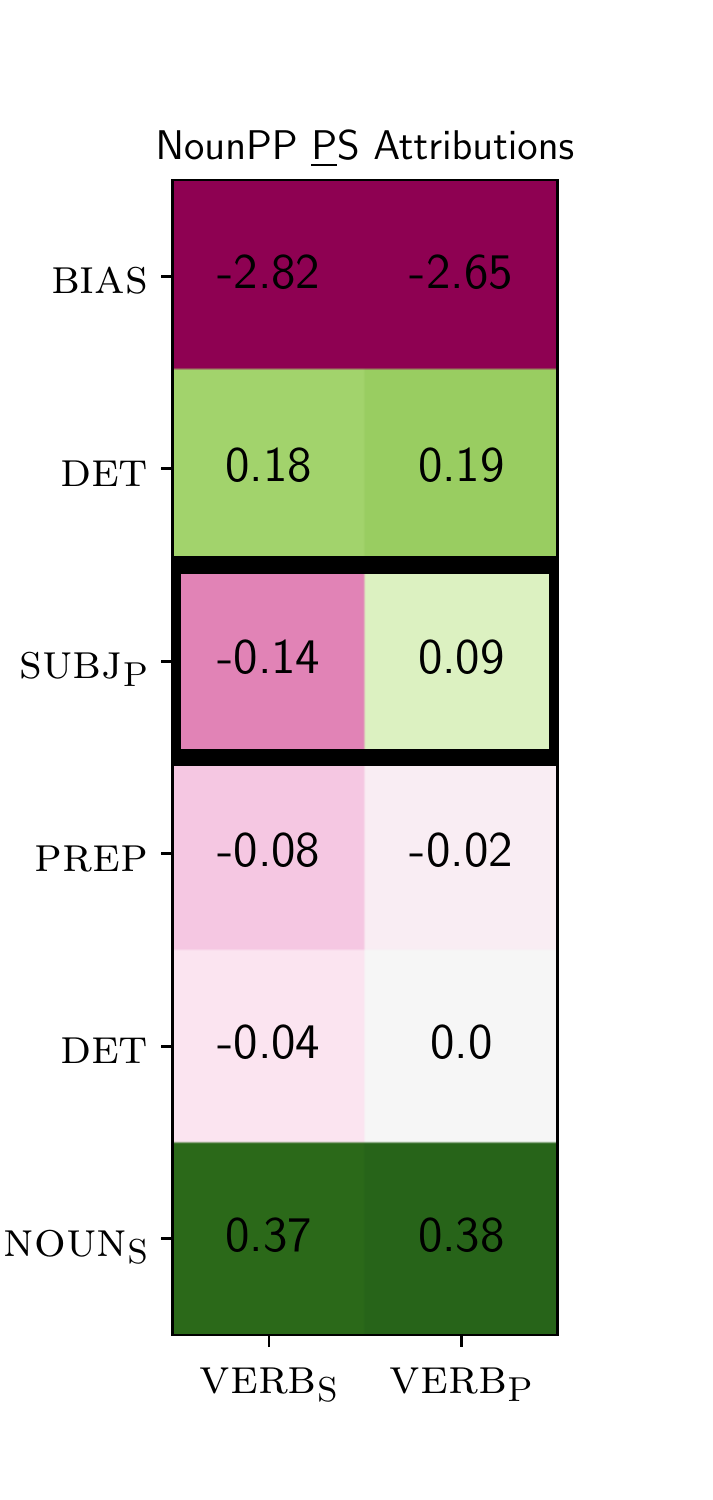}
            \label{fig:nounpp_ps_rnn}
        \end{subfigure}
        \caption{Aggregated LSTM attributions}
        \label{fig:nounpp_rnn}
    \end{subfigure}
    \caption{Contextual Decomposition attributions for the English models (SHA-RNN and LSTM) on the \underline{S}P and \underline{P}S conditions of the NounPP task. \cref{fig:nounpp_sharnn_single} shows the attributions of two individial sentences, while \cref{fig:nounpp_sharnn,fig:nounpp_rnn} show aggregated attributions over all sentences of that condition. Note that in \cref{fig:nounpp_sharnn} the attribution for the subject under the singular verb is both higher in the \underline{S}P condition as well as in \underline{P}S condition, while in \cref{fig:nounpp_rnn} the attribution is higher for the subject matching the verb form.}
    \label{fig:nounpp_attributions}
\end{figure*}

\subsubsection{Language axis}
Across the board, the Dutch models perform slightly better on the NA tasks than the English models. 
This could be due to the data sets used, as the Dutch data set was larger than the English one, giving the Dutch model more opportunities to learn. We do find similar patterns between the Dutch models (\cref{tab:pred_exp_dutch}) and the English models (\cref{tab:pred_exp_eng}): between the two languages, the models generally share the tasks and conditions that they perform well on. There are exceptions to this, as in the case of the Simple NA task for the LSTM, with Dutch models performing better on the singular condition while their English counterparts achieve higher scores on the plural condition. 

When we compare the results of the models on the Subject Attribution task in \cref{tab:pred_exp_dutch,tab:pred_exp_eng}, we find more substantial differences between the models across the languages. In case of the English models, the SHA-RNN performed rather poorly on the plural conditions of the Subject Attribution task. This is remarkable, given that the Dutch SHA-RNN yields significantly higher scores on these conditions.

We observe that for the English SHA-RNN, contextual decomposition consistently yields attribution scores  that are lower for the plural conditions than those for the singular conditions (see \cref{fig:nounpp_attributions} for an example). In the Dutch SHA-RNN, this behaviour is only apparent for the Simple, NounPP and NamePP tasks.

\citet{jumelet-etal-2019-analysing} encountered similar behaviours when applying CD to an LSTM language model. They attributed the lower attributions to a bias towards singular verbs in the model, which resulted in a form of default reasoning. However, our accuracy results do not indicate a similar bias, as we found all our models performing well on both plural and singular subjects.
This raises the question as to what is causing this behaviour, which we leave for future work.

Overall, these results do not demonstrate any significant differences between the Dutch and English models.
While we have shown that differences occur across conditions, we find that for most conditions, both models behave similarly, with the two LSTM models displaying more similarities than the SHA-RNN models.

\subsubsection{Attention axis}
To compare the attention models (SHA-RNNs) to the non-attention model (the LSTMs), we again first consider the accuracy scores in \cref{tab:pred_exp_dutch,tab:pred_exp_eng}. A comparison between the SHA-RNN and the LSTM shows that the SHA-RNN performs slightly worse than the LSTM by a small margin. There are some cases where this difference is more pronounced, such as for the English ThatNounPP task (see \cref{tab:pred_exp_eng}), where we observe large differences for the singular subject conditions. This behaviour goes against the perplexity results in \cref{tab:perplexity}, which indicate a better performing SHA-RNN. This is in line with the results found by \citet{DBLP:journals/corr/abs-2103-01819}, who demonstrate that perplexity is not always directly correlated to performance on downstream tasks, as appears to be the case for our Number Agreement task.

Looking at the model explanations in \cref{tab:pred_exp_dutch,tab:pred_exp_eng} we see that across the board the LSTM performs better on the Subject Attribution task. We find that both SHA-RNN models generally do not produce the expected attributions for the plural subject conditions, 
while there are very few instances of the LSTM performing under 50\%, only failing by a large margin for the English LSTM on the Simple \underline{P} and NamePP \underline{P} conditions (see \cref{tab:pred_exp_dutch}).

From our observations, the attention and non-attention models behave differently both in terms of accuracy scores on the NA task and the explanations from the Subject Attribution task. We find that the difference between the architectures of the SHA-RNN and the LSTM leads to significant variations in general performance as well as behavioural patterns.

\section{Conclusion}
In this paper, we compared both attention (SHA-RNN) and non-attention (LSTM) language models across two languages, namely Dutch and English. To test these models, we extended the Number Agreement task from \citet{lakretz-etal-2019-emergence} to the Dutch language, which allows us to compare these models across both languages. In addition to this, we extended a feature attribution method called Contextual Decomposition \cite{Murdoch2018BeyondLSTMs} to the SHA-RNN model. We applied Contextual Decomposition to the Number Agreement task to obtain interpretable explanations and compared the different models from a feature attribution standpoint.

We found that both the Dutch and English models behaved similarly in terms of accuracy. While general performance differed between the two languages, we did find that similar behavioural patterns emerged from the models. This partially held for the explanations obtained through Contextual Decomposition, where we did uncover differences. These differences were centred around the SHA-RNN, which we found behaved as if it applied default reasoning similar to the work of \citet{jumelet-etal-2019-analysing}. 

Comparing our attention and non-attention models, we found immediate differences, both when comparing the performance on the Number Agreement task as when looking into the attributions. 
Both models performed differently on the same tasks and feature attributions varied between them. 
We found that our LSTM performed better on the attribution task.

Our current results suggest that attention and non-attention models behave differently according to Contextual Decomposition. More specifically, we find that the attention models have more difficulty producing correct attributions for plural sentences. A logical next step would then be to compare our current results by those obtained through different attribution methods such as SHAP \cite{Lundberg2017APredictions} and Integrated Gradients \cite{Sundararajan2017AxiomaticNetworks}. 
Should we find that Contextual Decomposition holds up well to these other methods, it could then prove to be a valuable method for approximating Shapley values in polynomial time. Moreover, it is worth looking into the application of Contextual Decomposition in Transformer architectures, which rely more heavily on these kinds of attention mechanisms.

An alternative line of research that we would like to explore is the attention mechanism itself. Even though it has been shown that attention does not provide guarantees for explainability \citep{jain2019attention}, it would still be worthwhile to investigate the attention patterns that are employed by the SHA-RNN.

\bibliography{anthology,references_manual}

\begin{thebibliography}{36}
\expandafter\ifx\csname natexlab\endcsname\relax\def\natexlab#1{#1}\fi

\bibitem[{Ba et~al.(2016)Ba, Kiros, and Hinton}]{Ba2016LayerNormalization}
Lei~Jimmy Ba, Jamie~Ryan Kiros, and Geoffrey~E. Hinton. 2016.
\newblock \href {http://arxiv.org/abs/1607.06450} {Layer normalization}.
\newblock \emph{CoRR}, abs/1607.06450.

\bibitem[{Ettinger(2020)}]{ettinger2020bert}
Allyson Ettinger. 2020.
\newblock What bert is not: Lessons from a new suite of psycholinguistic
  diagnostics for language models.
\newblock \emph{Transactions of the Association for Computational Linguistics},
  8:34--48.

\bibitem[{Giulianelli et~al.(2018)Giulianelli, Harding, Mohnert, Hupkes, and
  Zuidema}]{giulianelli-etal-2018-hood}
Mario Giulianelli, Jack Harding, Florian Mohnert, Dieuwke Hupkes, and Willem
  Zuidema. 2018.
\newblock \href {https://doi.org/10.18653/v1/W18-5426} {Under the hood: Using
  diagnostic classifiers to investigate and improve how language models track
  agreement information}.
\newblock In \emph{Proceedings of the 2018 {EMNLP} Workshop {B}lackbox{NLP}:
  Analyzing and Interpreting Neural Networks for {NLP}}, pages 240--248,
  Brussels, Belgium. Association for Computational Linguistics.

\bibitem[{Goldberg(2019)}]{goldberga}
Yoav Goldberg. 2019.
\newblock \href {https://arxiv.org/pdf/1901.05287.pdf} {Assessing {{BERT}}'s
  {{Syntactic Abilities}}}.
\newblock page~4.

\bibitem[{Grave et~al.(2017)Grave, Joulin, and
  Usunier}]{Grave2016ImprovingCache}
Edouard Grave, Armand Joulin, and Nicolas Usunier. 2017.
\newblock \href {https://openreview.net/forum?id=B184E5qee} {Improving neural
  language models with a continuous cache}.
\newblock In \emph{5th International Conference on Learning Representations,
  {ICLR} 2017, Toulon, France, April 24-26, 2017, Conference Track
  Proceedings}. OpenReview.net.

\bibitem[{Gulordava et~al.(2018)Gulordava, Bojanowski, Grave, Linzen, and
  Baroni}]{Gulordava2018ColorlessHierarchically}
Kristina Gulordava, Piotr Bojanowski, Edouard Grave, Tal Linzen, and Marco
  Baroni. 2018.
\newblock \href {https://doi.org/10.18653/v1/n18-1108} {Colorless green
  recurrent networks dream hierarchically}.
\newblock In \emph{Proceedings of the 2018 Conference of the North American
  Chapter of the Association for Computational Linguistics: Human Language
  Technologies, {NAACL-HLT} 2018, New Orleans, Louisiana, USA, June 1-6, 2018,
  Volume 1 (Long Papers)}, pages 1195--1205. Association for Computational
  Linguistics.

\bibitem[{Hao(2020)}]{DBLP:journals/corr/abs-2005-00062}
Yiding Hao. 2020.
\newblock \href {http://arxiv.org/abs/2005.00062} {Attribution analysis of
  grammatical dependencies in lstms}.
\newblock \emph{CoRR}, abs/2005.00062.

\bibitem[{Hendrycks and Gimpel(2016)}]{Hendrycks2016GaussianGELUs}
Dan Hendrycks and Kevin Gimpel. 2016.
\newblock \href {http://arxiv.org/abs/1606.08415} {Bridging nonlinearities and
  stochastic regularizers with gaussian error linear units}.
\newblock \emph{CoRR}, abs/1606.08415.

\bibitem[{Hochreiter and Schmidhuber(1997)}]{Hochreiter1997LongMemory}
Sepp Hochreiter and Jürgen Schmidhuber. 1997.
\newblock \href {https://doi.org/10.1162/neco.1997.9.8.1735} {{Long Short-Term
  Memory}}.
\newblock \emph{Neural Computation}, 9(8):1735--1780.

\bibitem[{Hupkes et~al.(2018)Hupkes, Veldhoen, and
  Zuidema}]{hupkes2018visualisation}
Dieuwke Hupkes, Sara Veldhoen, and Willem Zuidema. 2018.
\newblock Visualisation and `diagnostic classifiers' reveal how recurrent and
  recursive neural networks process hierarchical structure.
\newblock \emph{Journal of Artificial Intelligence Research}, 61:907--926.

\bibitem[{Inan et~al.(2017)Inan, Khosravi, and Socher}]{Inan2016TyingModeling}
Hakan Inan, Khashayar Khosravi, and Richard Socher. 2017.
\newblock \href {https://openreview.net/forum?id=r1aPbsFle} {Tying word vectors
  and word classifiers: {A} loss framework for language modeling}.
\newblock In \emph{5th International Conference on Learning Representations,
  {ICLR} 2017, Toulon, France, April 24-26, 2017, Conference Track
  Proceedings}. OpenReview.net.

\bibitem[{Jain and Wallace(2019)}]{jain2019attention}
Sarthak Jain and Byron~C Wallace. 2019.
\newblock Attention is not explanation.
\newblock In \emph{Proceedings of the 2019 Conference of the North American
  Chapter of the Association for Computational Linguistics: Human Language
  Technologies, Volume 1 (Long and Short Papers)}, pages 3543--3556.

\bibitem[{Jumelet(2020)}]{Jumelet2020DiagNNose}
Jaap Jumelet. 2020.
\newblock diagnnose: A library for neural activation analysis.
\newblock In \emph{Proceedings of the Third BlackboxNLP Workshop on Analyzing
  and Interpreting Neural Networks for NLP}, pages 342--350.

\bibitem[{Jumelet and Hupkes(2018)}]{jumelet-hupkes-2018-language}
Jaap Jumelet and Dieuwke Hupkes. 2018.
\newblock \href {https://doi.org/10.18653/v1/W18-5424} {Do language models
  understand anything? on the ability of {LSTM}s to understand negative
  polarity items}.
\newblock In \emph{Proceedings of the 2018 {EMNLP} Workshop {B}lackbox{NLP}:
  Analyzing and Interpreting Neural Networks for {NLP}}, pages 222--231,
  Brussels, Belgium. Association for Computational Linguistics.

\bibitem[{Jumelet et~al.(2019)Jumelet, Zuidema, and
  Hupkes}]{jumelet-etal-2019-analysing}
Jaap Jumelet, Willem Zuidema, and Dieuwke Hupkes. 2019.
\newblock \href {https://doi.org/10.18653/v1/K19-1001} {Analysing neural
  language models: Contextual decomposition reveals default reasoning in number
  and gender assignment}.
\newblock In \emph{Proceedings of the 23rd Conference on Computational Natural
  Language Learning (CoNLL)}, pages 1--11, Hong Kong, China. Association for
  Computational Linguistics.

\bibitem[{Lakretz et~al.(2020)Lakretz, Hupkes, Vergallito, Marelli, Baroni, and
  Dehaene}]{DBLP:journals/corr/abs-2006-11098}
Yair Lakretz, Dieuwke Hupkes, Alessandra Vergallito, Marco Marelli, Marco
  Baroni, and Stanislas Dehaene. 2020.
\newblock \href {http://arxiv.org/abs/2006.11098} {Exploring processing of
  nested dependencies in neural-network language models and humans}.
\newblock \emph{CoRR}, abs/2006.11098.

\bibitem[{Lakretz et~al.(2019)Lakretz, Kruszewski, Desbordes, Hupkes, Dehaene,
  and Baroni}]{lakretz-etal-2019-emergence}
Yair Lakretz, German Kruszewski, Theo Desbordes, Dieuwke Hupkes, Stanislas
  Dehaene, and Marco Baroni. 2019.
\newblock \href {https://doi.org/10.18653/v1/N19-1002} {The emergence of number
  and syntax units in {LSTM} language models}.
\newblock In \emph{Proceedings of the 2019 Conference of the North {A}merican
  Chapter of the Association for Computational Linguistics: Human Language
  Technologies, Volume 1 (Long and Short Papers)}, pages 11--20, Minneapolis,
  Minnesota. Association for Computational Linguistics.

\bibitem[{Linzen et~al.(2019)Linzen, Chrupa{\l}a, Belinkov, and
  Hupkes}]{ws-2019-2019-acl}
Tal Linzen, Grzegorz Chrupa{\l}a, Yonatan Belinkov, and Dieuwke Hupkes,
  editors. 2019.
\newblock \href {https://www.aclweb.org/anthology/W19-4800} {\emph{Proceedings
  of the 2019 ACL Workshop BlackboxNLP: Analyzing and Interpreting Neural
  Networks for NLP}}. Association for Computational Linguistics, Florence,
  Italy.

\bibitem[{Linzen et~al.(2016)Linzen, Dupoux, and
  Goldberg}]{Linzen2016AssessingDependencies}
Tal Linzen, Emmanuel Dupoux, and Yoav Goldberg. 2016.
\newblock \href {https://transacl.org/ojs/index.php/tacl/article/view/972}
  {Assessing the ability of lstms to learn syntax-sensitive dependencies}.
\newblock \emph{Trans. Assoc. Comput. Linguistics}, 4:521--535.

\bibitem[{Lundberg and Lee(2017)}]{Lundberg2017APredictions}
Scott~M. Lundberg and Su{-}In Lee. 2017.
\newblock \href
  {https://proceedings.neurips.cc/paper/2017/hash/8a20a8621978632d76c43dfd28b67767-Abstract.html}
  {A unified approach to interpreting model predictions}.
\newblock In \emph{Advances in Neural Information Processing Systems 30: Annual
  Conference on Neural Information Processing Systems 2017, December 4-9, 2017,
  Long Beach, CA, {USA}}, pages 4765--4774.

\bibitem[{Merity(2019)}]{Merity2019SingleHead}
Stephen Merity. 2019.
\newblock \href {http://arxiv.org/abs/1911.11423} {Single headed attention
  {RNN:} stop thinking with your head}.
\newblock \emph{CoRR}, abs/1911.11423.

\bibitem[{Merity et~al.(2017)Merity, Xiong, Bradbury, and
  Socher}]{Merity2016PointerModels}
Stephen Merity, Caiming Xiong, James Bradbury, and Richard Socher. 2017.
\newblock \href {https://openreview.net/forum?id=Byj72udxe} {Pointer sentinel
  mixture models}.
\newblock In \emph{5th International Conference on Learning Representations,
  {ICLR} 2017, Toulon, France, April 24-26, 2017, Conference Track
  Proceedings}. OpenReview.net.

\bibitem[{Michel et~al.(2019)Michel, Levy, and Neubig}]{Michel2019AreOne}
Paul Michel, Omer Levy, and Graham Neubig. 2019.
\newblock \href
  {https://proceedings.neurips.cc/paper/2019/hash/2c601ad9d2ff9bc8b282670cdd54f69f-Abstract.html}
  {Are sixteen heads really better than one?}
\newblock In \emph{Advances in Neural Information Processing Systems 32: Annual
  Conference on Neural Information Processing Systems 2019, NeurIPS 2019,
  December 8-14, 2019, Vancouver, BC, Canada}, pages 14014--14024.

\bibitem[{Murdoch et~al.(2018)Murdoch, Liu, and Yu}]{Murdoch2018BeyondLSTMs}
W.~James Murdoch, Peter~J. Liu, and Bin Yu. 2018.
\newblock \href {https://openreview.net/forum?id=rkRwGg-0Z} {Beyond word
  importance: Contextual decomposition to extract interactions from lstms}.
\newblock In \emph{6th International Conference on Learning Representations,
  {ICLR} 2018, Vancouver, BC, Canada, April 30 - May 3, 2018, Conference Track
  Proceedings}. OpenReview.net.

\bibitem[{Nikoulina et~al.(2021)Nikoulina, Tezekbayev, Kozhakhmet, Babazhanova,
  Gall{\'{e}}, and Assylbekov}]{DBLP:journals/corr/abs-2103-01819}
Vassilina Nikoulina, Maxat Tezekbayev, Nuradil Kozhakhmet, Madina Babazhanova,
  Matthias Gall{\'{e}}, and Zhenisbek Assylbekov. 2021.
\newblock \href {http://arxiv.org/abs/2103.01819} {The rediscovery hypothesis:
  Language models need to meet linguistics}.
\newblock \emph{CoRR}, abs/2103.01819.

\bibitem[{Poerner et~al.(2018)Poerner, Sch{\"u}tze, and
  Roth}]{poerner2018evaluating}
Nina Poerner, Hinrich Sch{\"u}tze, and Benjamin Roth. 2018.
\newblock Evaluating neural network explanation methods using hybrid documents
  and morphosyntactic agreement.
\newblock In \emph{Proceedings of the 56th Annual Meeting of the Association
  for Computational Linguistics (Volume 1: Long Papers)}, pages 340--350.

\bibitem[{Ribeiro et~al.(2016)Ribeiro, Singh, and
  Guestrin}]{Ribeiro2016WhyClassifier}
Marco~T{\'{u}}lio Ribeiro, Sameer Singh, and Carlos Guestrin. 2016.
\newblock \href {https://doi.org/10.18653/v1/n16-3020} {"why should {I} trust
  you?": Explaining the predictions of any classifier}.
\newblock In \emph{Proceedings of the Demonstrations Session, {NAACL} {HLT}
  2016, The 2016 Conference of the North American Chapter of the Association
  for Computational Linguistics: Human Language Technologies, San Diego
  California, USA, June 12-17, 2016}, pages 97--101. The Association for
  Computational Linguistics.

\bibitem[{Samek et~al.(2019)Samek, Montavon, Vedaldi, Hansen, and
  M{\"u}ller}]{samek2019explainable}
Wojciech Samek, Gr{\'e}goire Montavon, Andrea Vedaldi, Lars~Kai Hansen, and
  Klaus-Robert M{\"u}ller. 2019.
\newblock \emph{Explainable AI: interpreting, explaining and visualizing deep
  learning}, volume 11700.
\newblock Springer Nature.

\bibitem[{Saphra and Lopez(2020)}]{saphra2020lstms}
Naomi Saphra and Adam Lopez. 2020.
\newblock Lstms compose—and learn—bottom-up.
\newblock In \emph{Proceedings of the 2020 Conference on Empirical Methods in
  Natural Language Processing: Findings}, pages 2797--2809.

\bibitem[{Shapley(1953)}]{shapley1953value}
Lloyd~S Shapley. 1953.
\newblock A value for n-person games.
\newblock \emph{Contributions to the Theory of Games}, 2(28):307--317.

\bibitem[{Simonyan et~al.(2014)Simonyan, Vedaldi, and
  Zisserman}]{DBLP:journals/corr/SimonyanVZ13}
Karen Simonyan, Andrea Vedaldi, and Andrew Zisserman. 2014.
\newblock \href {http://arxiv.org/abs/1312.6034} {Deep inside convolutional
  networks: Visualising image classification models and saliency maps}.
\newblock In \emph{2nd International Conference on Learning Representations,
  {ICLR} 2014, Banff, AB, Canada, April 14-16, 2014, Workshop Track
  Proceedings}.

\bibitem[{Sundararajan et~al.(2017)Sundararajan, Taly, and
  Yan}]{Sundararajan2017AxiomaticNetworks}
Mukund Sundararajan, Ankur Taly, and Qiqi Yan. 2017.
\newblock \href {http://proceedings.mlr.press/v70/sundararajan17a.html}
  {Axiomatic attribution for deep networks}.
\newblock In \emph{Proceedings of the 34th International Conference on Machine
  Learning, {ICML} 2017, Sydney, NSW, Australia, 6-11 August 2017}, volume~70
  of \emph{Proceedings of Machine Learning Research}, pages 3319--3328. {PMLR}.

\bibitem[{Tenney et~al.(2019)Tenney, Das, and Pavlick}]{tenney2019bert}
Ian Tenney, Dipanjan Das, and Ellie Pavlick. 2019.
\newblock Bert rediscovers the classical nlp pipeline.
\newblock In \emph{Proceedings of the 57th Annual Meeting of the Association
  for Computational Linguistics}, pages 4593--4601.

\bibitem[{Vaswani et~al.(2017)Vaswani, Shazeer, Parmar, Uszkoreit, Jones,
  Gomez, Kaiser, and Polosukhin}]{Vaswani2017AttentionNeed}
Ashish Vaswani, Noam Shazeer, Niki Parmar, Jakob Uszkoreit, Llion Jones,
  Aidan~N. Gomez, Lukasz Kaiser, and Illia Polosukhin. 2017.
\newblock \href
  {https://proceedings.neurips.cc/paper/2017/hash/3f5ee243547dee91fbd053c1c4a845aa-Abstract.html}
  {Attention is all you need}.
\newblock In \emph{Advances in Neural Information Processing Systems 30: Annual
  Conference on Neural Information Processing Systems 2017, December 4-9, 2017,
  Long Beach, CA, {USA}}, pages 5998--6008.

\bibitem[{Wolf(2019)}]{wolf2019}
Thomas Wolf. 2019.
\newblock \href {https://huggingface.co/bert-syntax/extending-bert-syntax.pdf}
  {Some additional experiments extending the tech report ''{{Assessing BERT}}'s
  {{Syntactic Abilities}}'' by {{Yoav Goldberg}}}.
\newblock page~7.

\bibitem[{You et~al.(2019)You, Li, Hseu, Song, Demmel, and
  Hsieh}]{DBLP:journals/corr/abs-1904-00962}
Yang You, Jing Li, Jonathan Hseu, Xiaodan Song, James Demmel, and Cho{-}Jui
  Hsieh. 2019.
\newblock \href {http://arxiv.org/abs/1904.00962} {Reducing {BERT} pre-training
  time from 3 days to 76 minutes}.
\newblock \emph{CoRR}, abs/1904.00962.

\end{thebibliography}
\bibliographystyle{acl_natbib}

\appendix

\end{document}